\documentclass[12pt]{elsarticle}
\usepackage{caption}
\usepackage{subcaption}

\usepackage{graphicx}
\usepackage[colorlinks=true,linkcolor=black, citecolor=blue, urlcolor=blue]{hyperref}
\usepackage{lineno}
\usepackage[utf8]{inputenc}
\usepackage{arabtex}
\usepackage{utf8}
\setcode{utf8}
\usepackage{color,soul}

\usepackage[labelsep=period]{caption}
\usepackage[labelfont=bf]{caption}
\def\endabstract{\egroup}
\journal{}

\usepackage{multirow}

\begin{document}
\makeatletter
\def\ps@pprintTitle{%
 \let\@oddhead\@empty
 \let\@evenhead\@empty
 \def\@oddfoot{}%
 \let\@evenfoot\@oddfoot}
\makeatother
\begin{frontmatter}

\title{Arabic aspect sentiment polarity classification using BERT}



\author[add1]{Mohammed M.Abdelgwad*}
\ead{mohammed.mustafa@aun.edu.eg}
\author[add1]{Taysir Hassan A Soliman}
\ead{taysirhs@aun.edu.eg}
\author[add1]{Ahmed I.Taloba}
\ead{Taloba@aun.edu.eg}

\address[add1]{
Information system department\\*
     Faculty of computer and information \\*
     Assiut university\\*
     Egypt}

\begin{abstract}
Aspect-based sentiment analysis(ABSA) is a textual analysis methodology that defines the polarity of opinions on certain aspects related to specific targets. The majority of research on ABSA is in English, with a small amount of work available in Arabic. Most previous Arabic research has relied on deep learning models that depend primarily on context-independent word embeddings (e.g.word2vec), where each word has a fixed representation independent of its context. This article explores the modeling capabilities of contextual embeddings from pre-trained language models, such as BERT, and making use of sentence pair input on Arabic aspect sentiment polarity classification task. In particular, we develop a simple but effective BERT-based neural baseline to handle this task. Our BERT architecture with a simple linear classification layer surpassed the state-of-the-art works, according to the experimental results on three different Arabic datasets. Achieving an accuracy of 89.51\% on the Arabic hotel reviews dataset, 73.23\% on the Human annotated book reviews dataset, and 85.73\% on the Arabic news dataset.
\end{abstract}

\end{frontmatter}

\section{Introduction}
ABSA is not a conventional sentiment analysis but a more difficult task, it is concerned with defining the aspect terms listed in a document, as well as the sentiments expressed against each aspect.\\
As demonstrated by \citep{b0}, Sentiment Analysis (SA) can be studied at three levels: the document level where the task is to identify sentiment polarities (positive, neutral, or negative) that is indicated throughout the entire document. The sentence level is concerned with classifying sentiments relevant to a single sentence. But the document contains many sentences and each sentence may contain multiple aspects with different sentiments, so the document and sentence level sentiment analysis may not be accurate and need another suitable type that makes this fine-grained analysis called ABSA.\\
ABSA was first launched on SemEval-2014 \citep{b1}, with the introduction of datasets containing annotated restaurant and laptop reviews. ABSA's work was largely replicated at SemEval over the next two years \citep{b2,b3} as the task has extended into various domains, languages, and challenges. SemEval-2016 provided 39 datasets in 7 domains and 8 languages for the ABSA task, additionally, the datasets were provided with Support Vector Machine (SVM) as a baseline evaluation procedure.\\
There are three primary tasks common to ABSA, as mentioned in \citep{b3}: aspect category identification (Task 1), aspect opinion target extraction (Task 2), and aspect polarity detection (Task 3). In this paper, we concentrate only on Task 3.\\
Neural Network (NN) variations have been applied to many Arabic Natural Language Processing (NLP) applications, such as Sentiment Analysis \citep{b35}, machine translation \citep{b36}, named entity recognition \citep{b37}, and speech recognition \citep{b38} and the highest results were obtained.\\
Using word embedding or distributed representations enhances neural network efficiency and improves the performance of Deep Learning (DL) models, therefore, it has been applied as a preliminary layer in various DL models.\\
Two types of word embeddings are available: contextualized and non contextualized word embeddings. Most of the research available on Arabic ABSA is based on non-contextualized word embedding, such as (word2vec and fastText). The main drawback of non-contextualized word embeddings is the presentation of a set of static word embeddings that do not take into account the various contexts in which they may appear.\\
In contrast, pre-trained language models based on transformers, such as BERT can provide dynamic embedding vectors which change by changing the context of words in the sentences. This made it more encouraging to use BERT in many tasks via fine-tuning on the downstream dataset related to the task.\\
Despite the fact that Arabic language has a significant number of speakers (estimated to be around 422 million \citep{b4}) and is a morphologically rich language, the number of studies in Arabic aspect-based sentiment analysis is still restricted due to its complexity.\\
The key contributions of the current work are as follows: 
\begin{itemize}
    \item
    
    To our knowledge, this is the first time a transfer learning-based model, such as BERT, has been used to handle Arabic Aspect sentiment polarity classification task.
   
    \item
    Unlike most current Arabic ABSA approaches, which rely largely on hand-crafted features and external resources, we present an end-to-end model for handling Aspect sentiment polarity classification task that do not need external resources or feature engineering efforts.
    
    \item
    Examining modeling competence of contextual embedding from pre-trained language models, such as BERT with sentence pair input on Arabic aspect sentiment classification task.
    \item
    A simple BERT based model with a linear classification layer was proposed to solve aspect sentiment polarity classification task. Experiments conducted on three different Arabic datasets demonstrated that, despite the simplicity of our model, it surpassed the state-of-the-art works.
   
\end{itemize}

The rest of the paper is organized as follows. Section 2 addresses the related work; Section 3 illustrates the proposed model;  Section 4 explains the datasets and the baseline procedures; Section 5 presents results and discussions; finally, section 6 concludes the paper.

\section{related work}
ABSA is an area of SA with research methodologies divided into two approaches: standard machine learning techniques and DL-based techniques.\\ 
ABSA's earliest efforts depended mainly on machine learning methods that focus on handcrafted features like lexicons to train sentiment classifiers \citep{b12}\citep{b13}. These methods are effective but rely heavily on the efficiency of handcrafted features.\\
Subsequently, a set of methods based on neural networks consisting of a word embedding layer followed by a neural architecture were developed for the ABSA task and pretty results were achieved \citep{b14,b15,b25}.\\
Several attention-based models have been applied to the ABSA task for its ability to focus on important parts of the sentence related to aspects \citep{b16} \citep{b17}. But a single attention network maybe not enough at capturing key dependency attributes between context and targets especially when the sentence is long, so multiple attention networks were proposed to solve this problem \citep{b20,b21}.\\
In order to resolve the question-answering problem, the authors of \citep{b42} have developed the memory network concept (MemNN), which was later adopted in several NLP challenges including ABSA\citep{b22,b23,b24}.\\
Pre-trained language models have recently played a vital role in many NLP applications, as they can take advantage of the vast volume of unlabeled data to learn general language representations; Elmo \citep{b27}, GPT \citep{b28}, and BERT \citep{b29} are among the most well-known examples.
The authors of \citep{b41} studied the use of BERT embeddings with several neural models, such as a linear layer, GRU model, self-attention network, and conditional random field to deal with the ABSA task.
In \citep{b40}, the authors have demonstrated that treating ABSA as a sentence pair classification task by building auxiliary sentence as input to the BERT, significantly improved the results.

\citep{gao2019target} created a straightforward architecture to evaluate the representational effectiveness of BERT in the ABSA task. Surprisingly, combining BERT with sophisticated neural networks—which previously performed well with embedding representations—does not consistently improve performance over that of the standard BERT-FC implementation. On the other hand, the addition of target data demonstrates steady accuracy enhancement.

\citep{song2019attentional} suggested attentional encoder network for the targeted sentiment classification task. (AEN) compares the text representation using pre-trained GloVe and BERT models and employs a multi-head attention (MHA) approach for encoding. The authors employed intra-MHA to encode the context and inter-MHA to encode the target word. Experiments and analysis show that the suggested model is effective and lightweight.
The strength of BERT embeddings in ABSA was further investigated by \citep{b30,b31,b32,b33}.\\
In general, Arabic ABSA research development is slower than English.

\citep{b8} combined aspect embedding with each word embedding and sent the mixture to CNN model to address aspect polarity and category identification tasks.
Subsequently, \citep{b10} took advantage of modeling internal information related to the hierarchical review structure in solving the ABSA task.

Two supervised machine learning-based techniques, RNN and SVM supplemented with a set of handcrafted features, were suggested by \citep{b34} and excellent results were achieved using SVM but RNN was faster in terms of training execution time.

To stimulate learning the connection between context words and targets, \citep{b7} combined aspect embedding with each word embedding and sent the mixture to LSTM. Then, the attention mechanism was applied to focus on context words related to certain aspects.

\citep{b45} proposed applying IAN network supported with Bi-GRU for extracting target and context representations in a better manner.\\
\citep{b9} proposed applying deep memory network based on a stack of IndyLSTM supplemented with recurrent attention network to solve aspect sentiment classification task.

\citep{mohammad2021gated} proposed investigating the use of the Multilingual Universal Sentence Encoder (MUSE) with the GRU model to improve previous outcomes for aspect extraction and aspect polarity classification tasks, in contrast to previous studies that used primarily word- or character-level representations, the MUSE model creates sentence-level embeddings.

\citep{bensoltane2022towards} attempted to investigate the modeling power of BERT in aspect-extraction and aspect-category identification tasks. In addition to examining the effects of adding stronger layers on top of BERT while dealing with the ATE task.

\citep{fadel2022arabic} proposed the BF-BiLSTM-CRF model based on BERT to handle the target extraction task. They combined contextualized string embedding with the BERT language model to improve word representation as embedding layer, and on top of it they stacked two Bilstm layers with crf layer as output layer.\\
In this paper, a simple BERT-based model with sentence pair input and a linear classification layer was proposed to handle the Arabic aspect sentiment classification task, and state-of-the-art results are achieved on three different Arabic datasets.

\section{BERT-based-model}
\begin{figure}[!ht]

 \includegraphics[width=1.2\textwidth]{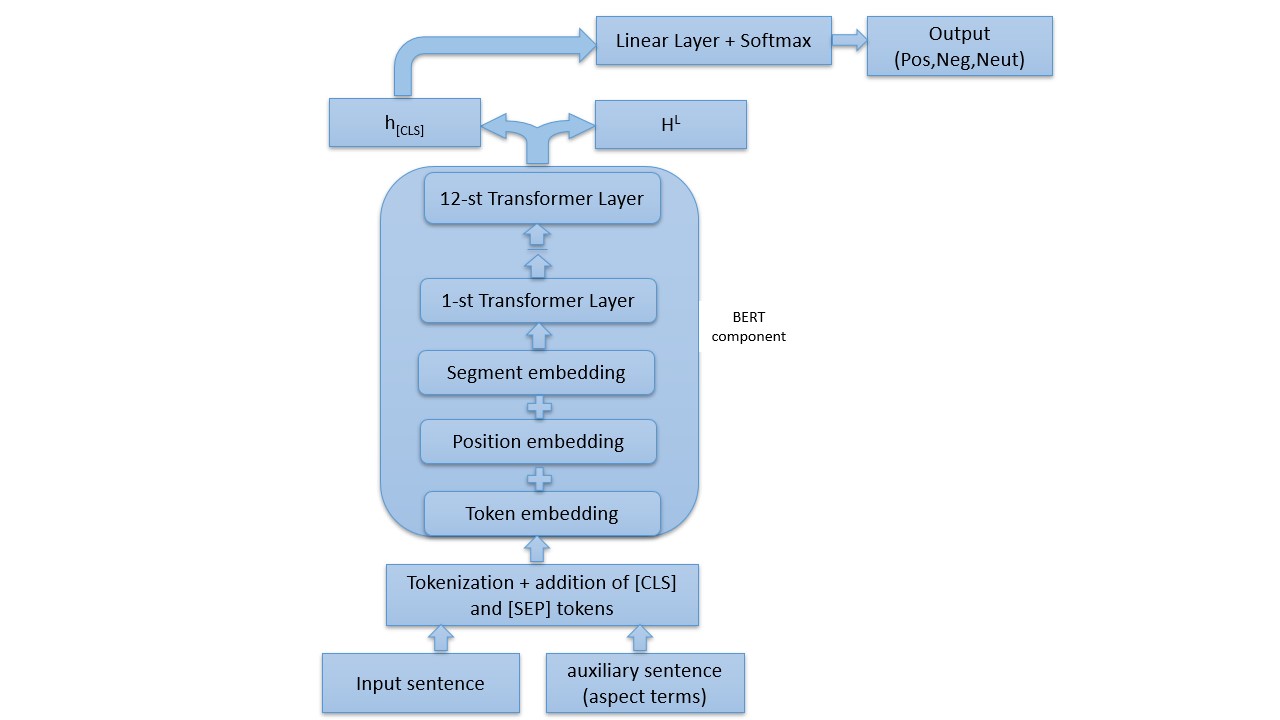}
    \caption{Model overall architecture.}
    \label{fig:Figure1}

\end{figure}

BERT (Bidirectional Encoder Representations from Transformers) is a deep learning technique for NLP in which deep neural networks use unsupervised language representation and bidirectional models built on Transformers (a deep learning algorithm in which each output element is linked to each input element and the weightings between them are dynamically determined based on their relation using the attention mechanism). BERT is pre-trained on two separate but related NLP tasks using the bidirectional capability: Masked Language Modeling and Next Sentence Prediction. BERT effectively handles ambiguity, which is the most difficult aspect of understanding natural language, and can reach high degree of accuracy in analyzing languages close to human beings.
We first use the BERT portion with L transformer layers to measure the corresponding contextualized representations for the T-length input token sequence. The hidden state representation of the [CLS] token is then fed to the task-specific layer to predict sentiment polarity labels.
The overall architecture of the proposed model is depicted in \autoref{fig:Figure1}.

\subsection{Auxiliary Sentence}
Since the BERT model accepts a single or pair of sentences as input, and due to the ability and effectiveness of the BERT model in dealing with sentence pair classification tasks, the ABSA task can be transformed into a sentence-pair classification task using the pre-trained BERT model, with the first sentence containing words that express sentiments within the sentence, and the second sentence containing information related to the aspect (the auxiliary sentence). In other words, the model receives the review sentence as the first sentence and the aspect terms as an auxiliary sentence, and the task would be to determine the sentiments towards each aspect.

\subsection{BERT as Embedding layer}
In comparison to the conventional word2vec embedding layer that offers static context-independent word vectors, the BERT layers offer dynamic context-dependent word embeddings by taking the entire sentence as input then calculating the representation of each word by extracting information from the entire sentence.\\
Inputs are processed in a special way by BERT, where sentences are tokenized first, as usual in every model, additionally, extra tokens are inserted at the start [CLS] and end [SEP] of the tokenized sentence. Then due to the utilization of self attention mechanism that enables BERT models to process tokens in parallel, and to deal with the next sentence prediction task, some special embedding tokens must be added to include all necessary information.\\
The tokenized sentence with [CLS] and [SEP] tokens are first fed into the embedding layer, which results in tokens embeddings. Those tokens embeddings don't include position information which added by means of position embeddings. Finally, it must be determined whether each token is associated with sentence A or B, this is possible by creating a new fixed token known as a segment embedding.\\
Then, the token embeddings, segment embeddings, and position embeddings for each token are combined and feed the mixture into $L$ transformer layers to optimize token level feature. The output representation $H^{l-1}$ from the $l-1$ transformer layer is fed as input into the next layer $l$ where $l\in[1,L]$.   
The representation $ H^{l} = \{h^{l}_1 ,...,h^{l}_t\}$ at the l-th layer is calculated as follows:

\begin{equation}
H^{l} = Transformer(H^{l-1})
\end{equation}
where $t$ refers to the number of input tokens.
Outputs from the last transformer layer $H^{L}$ are considered as full contextual representations for the input tokens. Then the last hidden state of BERT corresponding to [CLS] token $h_{[CLS]}$ is used as input to the task specific layer. 

\subsection{Design of Downstream Model}
In order to identify sentiment polarities toward aspects, word embeddings extracted from the BERT model are fed into a task-specific layer, a simple linear layer in our case, where both the input $h^l_t$ and the weight $W$ (the learnable parameters) matrices are multiplied with the addition of the bias term $b$ to transform their incoming features to output features in a linear manner. The softmax function is then used to determine the likelihood of each category $P$. 
\begin{equation}
P = softmax(W^Th^l_t + b)
\end{equation}

\section {Data and baseline research}
We conducted our experiments on 3 Arabic datasets: Human Annotated Arabic Dataset of Book Reviews (HAAD), the Arabic news dataset, and the Arabic Hotel Reviews Dataset.
The following subsections will describe each dataset in detail.

\subsection{ HAAD dataset}
\begin{figure}[!ht]
\begin{center}
 \includegraphics[width=1\textwidth]{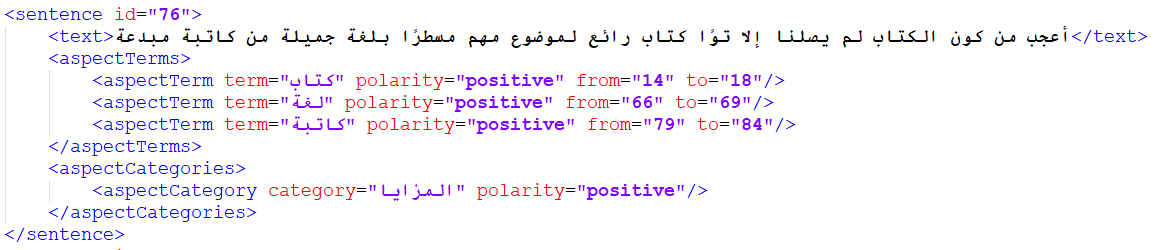}
    \caption{ Example of the HAAD Dataset Schema.}
    \label{fig:Figure2}
\end{center}
\end{figure}

For Arabic ABSA, the HAAD dataset \citep{b44} is regarded to be the first available dataset. There are 1513 Arabic book reviews in the HAAD dataset, each was annotated with aspect terms (T1), aspect term polarity (T2), aspect category (T3), and aspect category polarity (T4). This study focuses only on (T2). The SemEval-2014 framework was used to annotate the HAAD dataset. HAAD has a total of 2838 aspect terms, the distribution of which over the sentiment polarity classes (Positive, Negative, Conflict, Neutral) in both training and testing datasets are summarized in \autoref{tab:Tabel1}. The dataset was supported by baseline method for each task to compare with.\\
For the aspect term polarity baseline (T2): each aspect term in the test set was assigned the most common polarity label of that aspect in the training set, if it was found in the training set. If the aspect term was not in the training set, the most common label (Positive, Negative, Conflict, Neutral) in the training set will be used to label it in the test set.\\
\autoref{fig:Figure2} depicts an XML snapshot that corresponds to an annotated HAAD sentence.

\subsection{The Arabic news dataset}
\begin{figure}[!ht]
\begin{center}
 \includegraphics[width=1\textwidth]{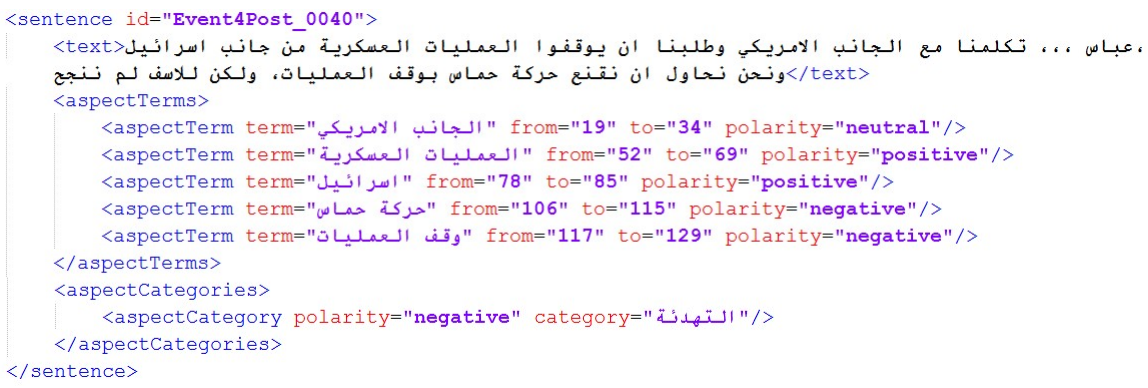}
    \caption{ Example of the Arabic news dataset Schema.}
    \label{fig:Figure3}
\end{center}
\end{figure}

The Arabic news dataset \citep{b43} comprises Facebook posts about the 2014 Gaza attacks and their comments. The SemEval-2014 framework was used to annotate the news dataset. There are 2265 news posts and 13628 comments in the Arabic news dataset each classified into three sentiment classes: Positive, negative, and neutral. Each post was manually annotated with aspect terms (T1), aspect term polarity (T2), aspect category (T3), and aspect category polarity (T4). Each comment was annotated with comment category detection (T5), and comment category polarity estimation (T6). This study focuses only on (T2). There are 9655 different aspects each related to one of four categories: Plans, Results, Peace, and Parties. \\
The dataset was supported by baseline method for each task to compare with.
For the aspect term polarity baseline (T2): 
the same baseline method previously applied in task 2 on the Human Annotated Arabic dataset of Book Reviews, was applied as a baseline procedure in this dataset.\\
\autoref{fig:Figure3} depicts an XML snapshot that corresponds to an annotated Arabic news sentence.

\begin{table}[!ht]
\caption{ Analysis of the HAAD dataset's aspect terms and polarities.}
\begin{center}
\footnotesize
\begin{tabular}{llllll}
\hline
\multirow{2}{*}{Dataset} & \multicolumn{4}{c}{Polarity} &  \multirow{2}{*}{Total} \\
& Positive&Negative&Conflict&Neutral& \\
\hline
Train& 1252 & 855 & 26 & 126 & 2259\\
\hline
Test& 124 & 432 & 1 & 22 & 579\\
\hline
Overall Dataset& 1376 & 1287 & 27 & 148 & 2838\\
\hline
\end{tabular}
\end{center}
\label{tab:Tabel1}
\end{table}

\subsection{The Arabic Hotel Reviews Dataset}
\begin{figure}[!ht]
\begin{center}
 \includegraphics[width=1\textwidth]{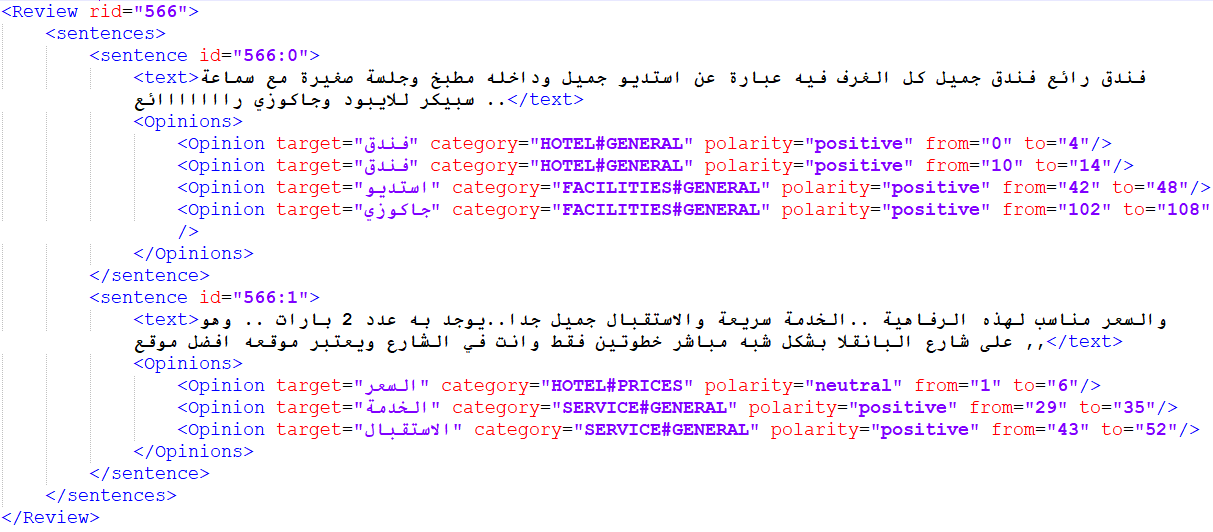}
    \caption{ Example of the Arabic Hotels Dataset Schema.}
    \label{fig:Figure4}
\end{center}
\end{figure}

The Arabic Hotel Reviews Dataset was presented in SemEval-2016 in support of ABSA's multilingual task involving work in 8 languages and 7 domains \citep{b3}. There are 19,226 training tuples and 4802 testing tuples in the dataset. 
The XML schema was used to annotate the dataset. 
The dataset consists of a set of reviews, each review contains a number of sentences, with each sentence having three tuples: aspect-category, OTE, and aspect polarity. \autoref{fig:Figure4} depicts an XML snapshot that corresponds to an annotated Arabic hotel review. 

The dataset supports both text-level annotations (2291 reviews) and sentence-level annotations (6029 sentences). This study focuses only on sentence-level tasks. The dataset's scale and distribution are explained in \autoref{tab:Tabel2}.
Also, SVM classifier supported with N-gram features was applied to the Arabic hotel review dataset for various ABSA tasks and was considered as baseline research to compare with. 

\begin{table}[!ht]
\caption{The scale and distribution of the Arabic hotels reviews dataset.}
\begin{center}
\footnotesize
\begin{tabular}{ccccccccc}
    \hline
    \multicolumn{3}{c}{TASK}&\multicolumn{3}{c}{TRAIN}&\multicolumn{3}{c}{TEST}\\
    & & &text&sentence&tuples&text&sentence&tuples\\
    \hline
    \multicolumn{3}{c}{T1: Sentence-level ABSA}&1839&4802&10.509&452&1227&2604\\
    \hline
    \multicolumn{3}{c}{T2: Text-level ABSA}&1839&4802&8757&452&1227&2158\\
    \hline
\end{tabular}
\end{center}
\label{tab:Tabel2}
\end{table}

\section{experiments}
Our models was trained and tested separately on 3 different Arabic datasets. When training the models on each dataset, 70 \% of the dataset was used, for validation 10\% was used and for testing 20\%. The Pytorch library was used to implement all the neural networks. All models computations were carried out independently on the GeForce GTX 1080 Ti GPU. The following subsections explain model training in detail.

\subsection{Evaluation method}
In order to determine the effectiveness of the proposed model, the accuracy metric was adopted, which was defined as follows:
\begin{equation}
Accuracy = \frac{correct\_predictions\_number}{overall\_samples\_number}
\end{equation}
Accuracy measures the number of correct samples to all samples, higher accuracy indicates better performance.

\subsection{Pre-processing}
We evaluated our BERT-based models with various preprocessing steps such as stemming, normalization, and stop words removal and discovered that there was no improvement in the outcomes, but rather that they were worse. This can be explained by the fact that these processes have a negative impact on the contextual meaning of the words learnt using the BERT. Consequently, we chose not to apply any of these procedures to the datasets.

\subsection{Hyperparameters Setting}
The pretrained "Arabic BERT" \citep{b39} was used, which was previously trained on about 8.2 billion words of MSA and dialectical Arabic. The BERT-Base model consisting of 12 hidden layers, 12 attention heads, and hidden size of 768 has been particularly used. 
Adam optimizer was used to fine-tune the model on the downstream task with a learning rate of 1e-5, dropout rate of (0.1 for Arabic hotel reviews dataset and 0.3 for both HAAD and Arabic news datasets), hidden dropout probability of 0.3, batch size of (24 for Arabic hotel reviews dataset, 16 for HAAD and 64 for Arabic news dataset), and number of epochs equal to 10.

\subsection{comparison models}
\noindent
\textbf{LSTM} just one LSTM is used for sentence modeling, with the last hidden states used as a representation for final classification.\\
\textbf{TD-LSTM} it employs two LSTM networks to model the target's prior and subsequent contexts to provide target-dependent representation for sentiment prediction \citep{tang2015effective}.\\
\textbf{INSIGHT-1} combined aspect embedding with each word embedding and fed the resulting mixture to CNN for Aspect sentiment analysis \citep{b8}. \\
\textbf{HB-LSTM} developed hierarchical bidirectional LSTM for ABSA, that can take advantage of hierarchical modeling information of the review in improving performance \citep{b10}.\\
\textbf{AB-LSTM-PC}  combined aspect embedding with each word embeddings to motivate learning the connections between context words and targets, then applied the attention mechanism for focusing on context words related to specific aspects \citep{b7}.\\
\textbf{IAN-BGRU} Used Bi-GRU to extract hidden representations from targets and context, then applied two associated attention networks on those representations to model targets and their context in an interactive manner \citep{b6}.\\
\textbf{MBRA}  made use of external memory network containing a stack of bidirectional lndy-lstms consisted of 3 layers, and a recurrent attention mechanism to deal with complex sentence structures \citep{b9}.

\subsection{Results and Discussion}
\autoref{tab:Tabel3} shows that simply adding a basic linear layer on top of BERT, outperformed the baselines and achieved better results than many previous Arabic DL models. This may be justified by the superior ability of the BERT model to extract semantic representations compared to the context-free word embedding models. In particular, during the training phase, BERT can learn information from both directions and is more adept at dealing with the OOV problem than other context-free word embedding models. Moreover, the use of the auxiliary sentence further improved the results of the BERT model, which is apparent in the higher results of the BERT-pair model compared to the BERT-single, achieving state-of-the-art results.This is evidence of the effectiveness of Bert's contextual representations at encoding associations between aspect terms and context words.

\begin{table}[!ht]
\caption{ Models Accuracy Results on the Arabic hotel reviews dataset, Arabic news dataset, and HAAD}
\label{tab:Tabel3}
\begin{center}
\begin{tabular}{ c  c  c  c }
\hline
\textbf{Models} & \multicolumn{3}{ c }{\textbf{Datasets}}  \\ 
\cline{2-4}

& \textbf{Arabic hotel reviews} & \textbf{Arabic news} & \textbf{HAAD} \\
\hline

Baseline &  76.4 & 61.47  & 29.70 \\ \hline
LSTM & 81.49 & 74.81 & 52.30 \\ \hline
TD-LSTM & 81.79  & 76.62 & 54.26 \\ \hline
AB-LSTM-PC  & 82.60 & 77.11 & 58.08 \\ \hline
INSIGHT-1 (CNN) & 82.71 & -  & - \\ \hline
HP-LSTM &  82.80& - & -\\ \hline
IAN-BGRU &  83.98& 78.78  & 68.62\\ \hline
MBRA &  87.31 & -  & - \\ \hline
BERT-Linear-single & 85.93 & 82.12  & 68.25  \\ \hline
\textbf{BERT-Linear-pair} & \textbf{89.51} &  \textbf{85.73} & \textbf{73.23} \\ \hline
\end{tabular}
\end{center}
\end{table}

\subsubsection{Overfitting Issue}
\begin{figure}
     \centering
     \begin{subfigure}[b]{0.6\textwidth}
         \centering
         \includegraphics[width=\textwidth]{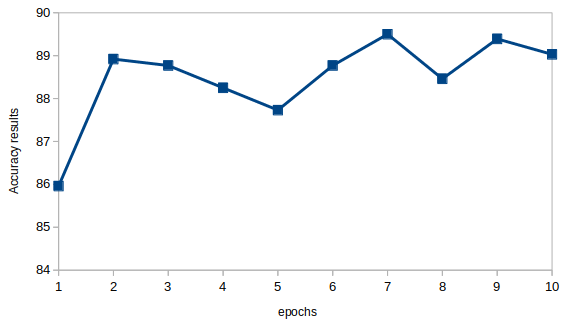}
         \caption{Performances on the Arabic hotels reviews dataset.}
         \label{fig:Figure5.1}
     \end{subfigure}
     \hfill
     \begin{subfigure}[b]{0.6\textwidth}
         \centering
         \includegraphics[width=\textwidth]{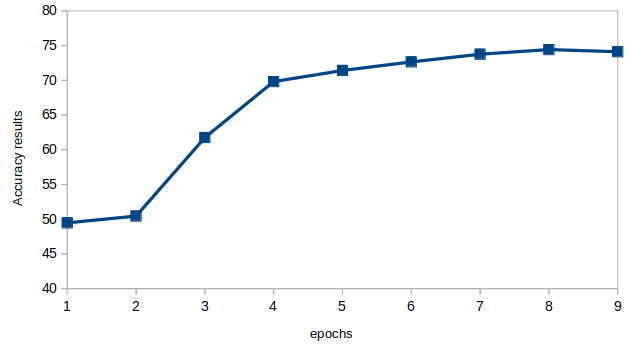}
         \caption{Performances on the Arabic news dataset.}
         \label{fig:Figure5.2}
     \end{subfigure}
     \hfill
     \begin{subfigure}[b]{0.6\textwidth}
         \centering
         \includegraphics[width=\textwidth]{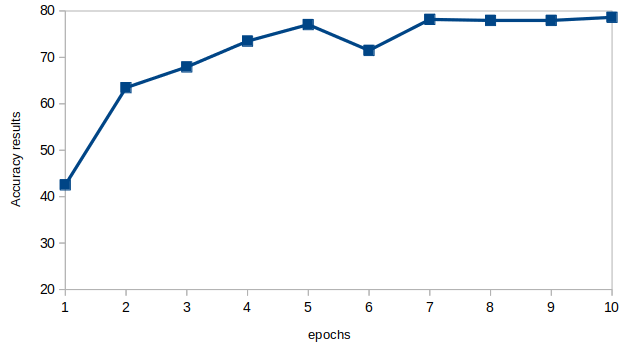}
         \caption{Performances on the HAAD.}
         \label{fig:Figure5.3}
     \end{subfigure}
        \caption{The accuracy results on the development set for three different Arabic datasets.}
        \label{fig:Figure5}
\end{figure}

Despite the use of Bert-base model "the smallest pretrained version of Bert", the number of parameters seemed to be large (110M) for this task, which made us wonder: Is our model overfitting the downstream data ? so, we trained the BERT-linear model on three different Arabic datasets for 10 epochs and noticed the oscillating accuracy results on the development sets after each epoch. As indicated in \autoref{fig:Figure5}, the accuracy results of the development sets are relatively stable and do not decrease significantly as the training progresses, which reveals that the BERT model is extremely robust to overfitting.

\subsubsection{finetuning or not}
\begin{figure}[!ht]
\begin{center}
 \includegraphics[width=.7\textwidth]{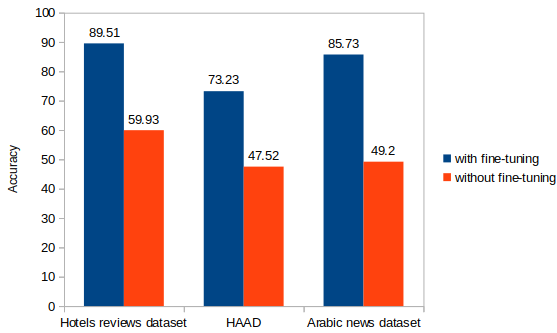}
    \caption{Effect of fine-tuning BERT on different datasets.}
    \label{fig:Figure6}
\end{center}
\end{figure}

We investigated the effect of fine-tuning on the final results by keeping the parameters of the BERT component fixed during the training phase. \autoref{fig:Figure6} shows a simple comparison between the performance of the model when fine-tuning and when setting the parameters fixed. The general purpose BERT representation is obviously far from acceptable for the downstream task, and task-specific fine-tunning is necessary to use BERT's capabilities to increase performance.

\subsection{Case Study}
The assessment results demonstrated the superiority of our model over traditional deep learning models that employ context-free word embeddings and feature-based machine learning techniques.

\autoref{tab:Tabel4} shows the ability of BERT to detect true polarity labels when the sentence contains one or more targets with the same polarity. As shown in this example "The furniture is very old and the service is poor and I would not recommend anyone to stay there", there are two aspects (furniture,service) with the same polarity (Neg,Neg), BERT models could identify the negative polarity toward both aspects.

Furthermore, by including target information as an auxiliary sentence, our BERT-pair model was able to determine the true polarity label for each target, particularly when the sentence has many targets with conflict polarities.
For example, in this review sentence "The service is bad but the hotel is very nice and beautiful", there are two different aspects (Service,Hotel) with two different sentiments (Neg and Pos). Despite this challenge, our model is capable of accurately identifying and determining the expressed polarity toward each aspect.

The ability to determine sentiment polarity when the sentence contains shifting words (e.g. not) is one of the most common problems faced by most SA algorithms. The issue is that, shifting words cause the sentiment polarity of a review or aspect to change completely. 
To properly classify sentiment polarity when the sentence contains shifting words, dedicated dictionaries to these words were used, as in most well-known SA methods, but our model is capable of determining this without the use of any other resources, such as lexicons.
For example, in this sentence “The hotel location is not good for the elderly” our model can easily determine the sentiment polarity toward (Hotel\#Location) as Negative.

But employing negation does not always result in a change in the sentiment polarity. For instance, our model is unaffected by the negation and can classify the aspect Hotel\#Location into Positive sentiment polarity in the review sentence “The hotel location is not only suitable for elderly people, but also is very close to most of the city historical sights”.
This is evidence of the effectiveness of Bert's contextual representations at encoding associations between targets' and context words.

\begin{table}[tbp]
\centering
\resizebox{\columnwidth}{!}{%
\begin{tabular}{|l|l|l|l|l|}
\hline
\multicolumn{1}{|c|}{\textbf{Sentence}} &
  \multicolumn{1}{c|}{\textbf{Target}} &
  \multicolumn{1}{c|}{\textbf{Bert-single}} &
  \multicolumn{1}{c|}{\textbf{Bert-pair}} &
  \multicolumn{1}{c|}{\textbf{True label}} \\ \hline
\begin{tabular}[c]{@{}l@{}}
Ar: \RL{فندق لطيف للغاية ، جميل ، عائلي ومناسب}\\
Eng: A very sweet, beautiful, family and suitable hotel\end{tabular} & \RL{فندق}  & Positive    & Positive  & Positive  \\ \hline
\begin{tabular}[c]{@{}l@{}}
Ar: \RL{موقع الفندق غير مناسب لكبار السن}\\
Eng: The hotel location is not good for the elderly \end{tabular}    & \RL{موقع الفندق} & Negative    & Negative  & Negative \\ \hline
\begin{tabular}[c]{@{}l@{}}

Ar: \RL{موقع الفندق ليس مناسبًا لكبار السن فحسب ، ولكنه قريب جدًا أيضًا من معظم مناطق الجذب التاريخية في المدينة}\\

Eng: The hotel location is not only suitable for elderly people,\\ but also is very close to most of the city historical sights\end{tabular} &
  \RL{موقع الفندق} &
  Positive &
  Positive &
  Positive \\ \hline
\multirow{2}{*}{\begin{tabular}[c]{@{}l@{}}
Ar: \RL{الأثاث قديم جدًا والخدمة سيئة ولا أنصح أي شخص بالبقاء هناك
} \\
Eng: The furniture is very old and the service is poor and I would not advise anyone to stay there\end{tabular}} &
  \RL{الأثاث} &
  Negative &
  Negative &
  Negative \\ \cline{2-5} 
                                                   & \RL{الخدمة}        & Negative    & Negative  & Negative \\ &&&&\\\hline
\multirow{3}{*}{\begin{tabular}[c]{@{}l@{}}
Ar: \RL{الخدمة سيئة ولكن الفندق رائع وجميل جدا}\\
Eng: The service is bad but the hotel is great and very beautiful\end{tabular}} &
  \RL{الخدمة} &
  Positive &
  Negative &
  Negative \\ \cline{2-5} 
                                                   & \RL{الفندق}          & Positive    & Positive  & Positive \\ &&&&\\ \hline 
\end{tabular}%
}
\caption{Case study of different models on the Arabic hotel reviews dataset}
\label{tab:Tabel4}
\end{table}

\section{Conclusion and future work}
In this paper, we explored the modeling capabilities of contextual embeddings from the pre-trained BERT model with the benefit of sentence pair input on the Arabic aspect sentiment polarity classification task. specifically, we examined the incorporation of the BERT embedding component with a simple linear classification layer and extensive experiments were performed on three Arabic datasets. The experimental results show that despite the simplicity of our model, it surpassed the state-of-the-art works, and is robust to overfitting.

For future work, we intend to enhance the BERT-linear-pair model by replacing the linear layer on top of the BERT embedding layer, with recurrent neural networks, self-attention networks or other sophisticated networks such as AEN. We also plan to use transformer-based models in other Arabic ABSA tasks such as aspect extraction and aspect category detection.

\section{Declarations}

\subsection{Ethics approval and consent to participate
}
Not Applicable.
\subsection{Consent for publication
}
Not applicable
\subsection{Competing interests}
There is no conflict of interest

\subsection{Availability of data and materials
}
The datasets analyzed during the current study are available in :\\
HAAD: \url{https://github.com/msmadi/HAAD}\\
Arabic hotel reviews dataset: \url{https://github.com/msmadi/ABSA-Hotels}\\
Arabic news dataset: \url{https://github.com/malayyoub/ABSA-for-Affective-News-Analysis-}

\subsection{Funding details}
This research did not receive any specific grant from funding agencies in the public, commercial, or not-for-profit sectors.

\subsection{Acknowledgment}
Not Applicable.

\subsection{Author contributions}
\textbf{Mohammed M.Abdelgwad}: Have done all the work in the paper.
\textbf{Taysir Hassan A Soliman}: Review \& editing, Supervision. \textbf{Ahmed I.Taloba}: Review \& editing.

\bibliography{sample.bib}
\typeout{get arXiv to do 4 passes: Label(s) may have changed. Rerun}
\end{document}